\documentclass[11pt]{article}
\usepackage{coling2020}
\usepackage{times}
\usepackage{url}
\usepackage{latexsym}

\usepackage{microtype}

\usepackage{mathtools}

\usepackage{tabularx}

\usepackage[utf8]{inputenc}
\usepackage{fourier} 
\usepackage{makecell}
\usepackage{multirow}
\usepackage{skt}

\usepackage{devanagari}  
\usepackage[utf8]{inputenc}

\usepackage{subfig}

\colingfinalcopy 

\title{Joint Transformer/RNN Architecture for \\Gesture Typing in Indic Languages}

\author{Emil Biju, Anirudh Sriram, Mitesh M. Khapra, Pratyush Kumar \\
  Indian Institute of Technology Madras \\
  \texttt{\{ee17b071,ee18b073\}@smail.iitm.ac.in} \\
  \texttt{miteshk@cse.iitm.ac.in} \\
  \texttt{pratyushkpanda@gmail.com}}

\date{\today}

\begin{document}
\maketitle
\begin{abstract}
Gesture typing is a method of typing words on a touch-based keyboard by creating a continuous trace passing through the relevant keys. This work is aimed at developing a keyboard that supports gesture typing in Indic languages. We begin by noting that when dealing with Indic languages, one needs to cater to two different sets of users: (i) users who prefer to type in the native Indic script (Devanagari, Bengali, etc.) and (ii) users who prefer to type in the English script but want the transliterated output in the native script. In both cases, we need a model that takes a trace as input and maps it to the intended word. To enable the development of these models, we create and release two datasets. First, we create a dataset containing keyboard traces for 193,658 words from 7 Indic languages. Second, we curate 104,412 English-Indic transliteration pairs from Wikidata across these languages. Using these datasets we build a model that performs path decoding, transliteration and transliteration correction. Unlike prior approaches, our proposed model does not make co-character independence assumptions during decoding. The overall accuracy of our model across the 7 languages varies from 70-95\%. 
\end{abstract}

\section{Introduction}
\label{intro}
\blfootnote{
    \hspace{-0.52cm}  
    This work is licensed under a Creative Commons 
    Attribution 4.0 International Licence.
    Licence details:
    \url{http://creativecommons.org/licenses/by/4.0/}.}

    %
    %


Gesture typing \cite{yang2019investigating} is a fast and convenient method for providing textual input to a touch-based keyboard on small-sized devices like mobile phones. The conventional method of prodding the keys to input a word requires several discrete key presses resulting in impeded performance \cite{zhai2012word}. Besides, touch-based keyboards do not provide the rich tactile sensation of key boundaries and key presses that mitigates erroneous typing on a regular desktop keyboard \cite{kristensson2007discrete}. Gesture typing allows the user to input a word by drawing a single continuous trace over the keyboard and the finger needs to be lifted only once a word is completed. It takes advantage of muscle memory \cite{yang2019investigating} which allows the user to subconsciously memorise input shapes for common words, resulting in improved typing speed and accuracy. 
   
The advantages of gesture typing over conventional typing methods are of particular significance in Indic language keyboards as Indic languages have a larger character set than English (for example,the Hindi script contains 52 alphabets as opposed to 26 in English), thus making the keyboard denser and increasing the chances of prodding the wrong character. A solution to this is to use an English character keyboard which receives an input sequence of English characters phonetically resembling the Indic word and transliterates them into the Indic language word \cite{hellsten2017transliterated}. Besides, prior experience with using English keyboards ensures that minimal effort is elicited from the user while switching to the new keyboard. Therefore, in this work, we propose solutions to two tasks: (i) \textbf{English-to-Indic Decoding} to predict an Indic word from gesture input provided to an English character keyboard with phonetic correspondence to the intended Indic word and (ii) \textbf{Indic-to-Indic Decoding} to predict an Indic word from the gesture input provided to an Indic character keyboard. To enable development on both these fronts, we release datasets of word traces simulated using the minimum jerk principle \cite{Quinn2018ModelingGM}. 

\begin{figure}[!t]
    \centering
   	\subfloat{\includegraphics[scale=0.15]{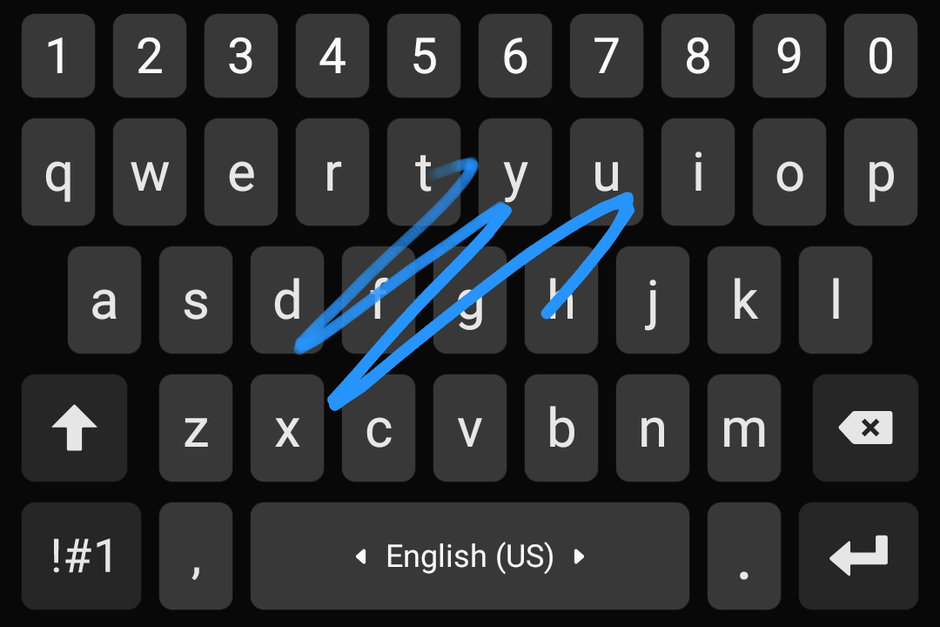}}%
   	\qquad
   	\qquad
   	\subfloat{\includegraphics[scale=0.30]{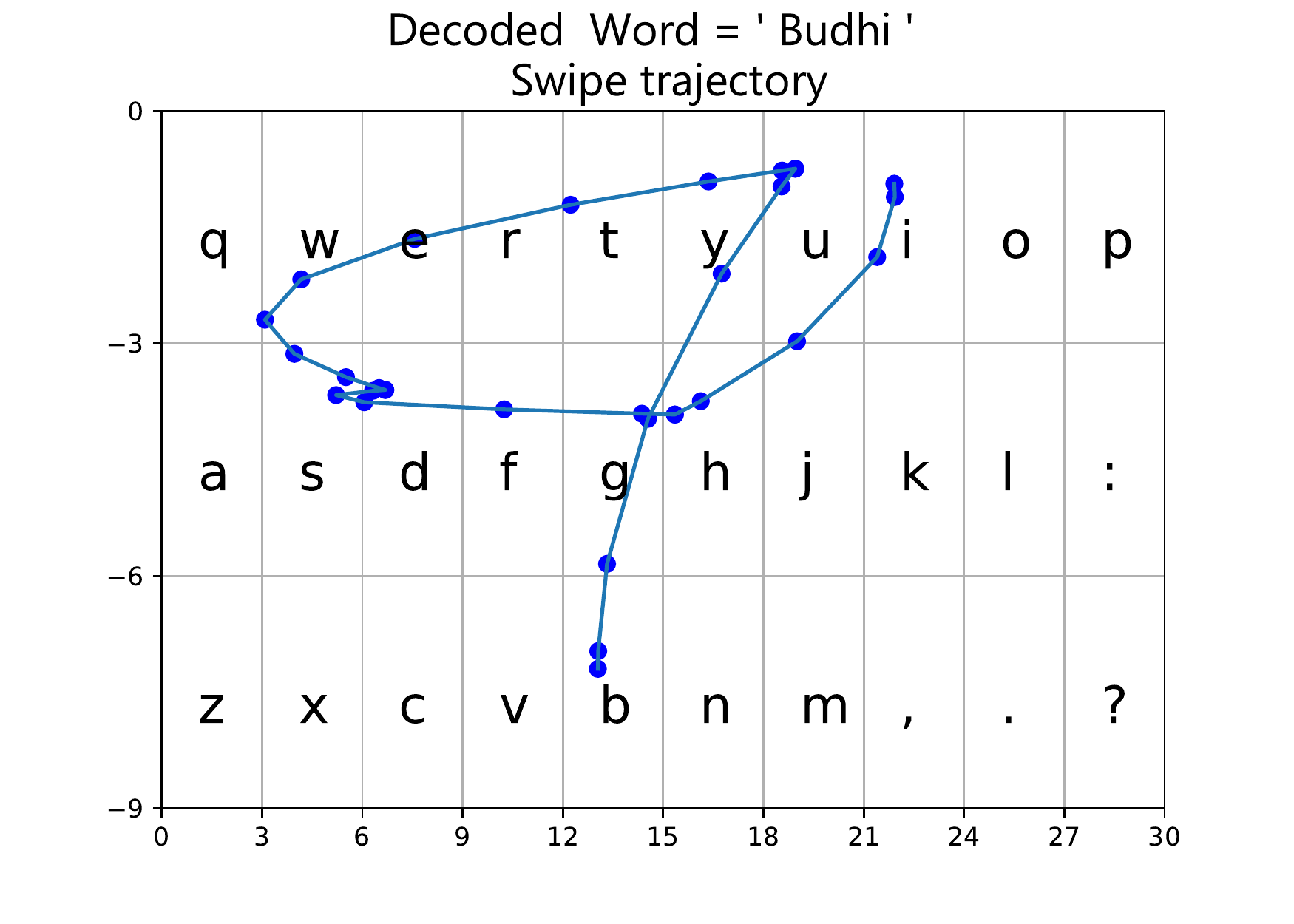}}%
   	\caption{(a) Gesture typing on a smartphone English script keyboard (b) A sample trace for the word \textit{budhi} that was simulated using the minimum jerk principle}
   	\label{fig:gesture_sample}
   \end{figure} 
   

In this paper, we propose to jointly use a multi-headed Transformer encoder layer \cite{vaswani2017attention} and Bidirectional LSTM layers to encode a sequential stream of $(x,y)$ coordinates obtained from a trace drawn on a touch-based keyboard. We decode the intended character sequence using an LSTM decoder to avoid any co-character independence assumptions. We use an LSTM encoder-decoder framework with attention for transliterating the decoded sequence into the intended Indic language. We also propose a neural network architecture for spelling correction of generated transliterations that uses character embeddings generated using the ELMo neural network \cite{peters-etal-2018-deep} to measure proximity between words. Unlike prior work, this model requires a smaller training set and allows easy vocabulary augmentation without model re-training. To sum up, our novel contributions include:
\begin{itemize}
    \item A Gesture Path Decoding model that uses a multi-headed Transformer along with LSTM layers for coordinate sequence encoding and a character-level LSTM model for character sequence decoding.
    \item A Contrastive Transliteration correction model that uses position-aware character embeddings to measure word proximities and correct spellings of transliterated words.
    \item Two datasets of simulated word traces for supporting work on gesture typing for Indic language keyboards including low resource languages like Telugu and Kannada.
    \item The accuracies of the proposed models vary from 70 to 89\% for English-to-Indic decoding and 86-95\% for Indic-to-Indic decoding across the 7 languages used for the study.
\end{itemize}

\section{Related Work}
\subsection{Gesture Recognition}
Classical techniques for gesture recognition include region encoding \cite{teitelman1964real}, geometric feature matching, linear machine classifiers \cite{rubine1991specifying}, Hidden Markov Models \cite{wilson2001hidden} and Dynamic time warping \cite{bautista2012probability}. Kristensson \shortcite{kristensson2007discrete} proposes to jointly use a shape channel and a location channel to make a prediction by comparisons based on both the shape and location of the gesture input. Improvements in gesture typing using LSTMs have been proposed which uses the CTC loss function for training \cite{alsharif2015long}. CTC uses dynamic programming to efficiently compute probabilities of all frame-level alignments that predict the target sequence and maximises their sum. Stimulated learning for training the CTC model to encode dependence on previously predicted words has also been proposed \cite{heymann2019improving}. However, these approaches assume that the character predicted at each step is independent of other characters for a given word.  

\subsection{Speech Recognition}
Speech recognition 
shares strong conceptual similarities with gesture input decoding. RNN-transducers which use CTC-based sequence encoding along with a word-level language model have also been used for speech recognition \cite{he2019streaming}. The work in \cite{chan2016listen} proposes a pyramidal recurrent network to encode the speech signal and an attention-based recurrent network decoder to predict the characters. Speech recognition has also been performed using a hybrid attention model \cite{chorowski2015attention} that uses both context and location-based features for character prediction. However, large variations between the input and output lengths in gesture input decoding make it difficult to track the alignment in attention-based models.

\subsection{Spelling Correction and Language Models}
Classical approaches to spelling correction include probabilistic modelling \cite{church1991probability,goodman2002language}, geometric pattern matching \cite{10.1145/1040830.1040867}, n-gram models \cite{ahmed2009revised} and edit distance-based models\cite{wagner1974string}. The work by Kim et al. \shortcite{kim2016character} a CNN (Convolutional Neural Network) to extract character level features which are then run through a highway network, followed by a word-level Recurrent Neural Network to predict the next word in the sentence. The work by Ghosh and Kristensson \shortcite{ghosh2017neural} extracts character level features by jointly using a GRU and CNN and generates a linear transformation of the outputs of the two networks. A decoder uses these features to predict the most probable word. However, this model uses a small-sized vocabulary and does not use position-aware character embeddings. Augmenting the vocabulary of most neural network approaches requires a new round of training as the decoder predictions are based only on previously seen samples and does not rely on any proximity metrics between the misspelt and vocabulary words.

\section{Gesture Data Generation} \label{sec: data_gen}
For this work, we curated two datasets\footnote{Our datasets and code are available at \url{https://iitmnlp.github.io/indic-swipe/}} for performing the Indic-to-Indic decoding and English-to-Indic decoding tasks across the 7 Indic languages used in the study. The first dataset contains Indic words obtained from the work by Goldhahn et al. \shortcite{goldhahn2012building} along with corresponding coordinate sequences depicting the path traced on an Indic character smartphone keyboard to input the word. The second dataset contains Indic-English transliteration pairs scraped from Wikidata along with corresponding coordinate sequences for the path traced on an English character keyboard. We use the principle that human motor control tends to maximise smoothness in movement to develop a synthetic path generation technique that models the gesture typing trajectory as a path that minimises the total jerk \cite{Quinn2018ModelingGM}.

For simulating the path between two characters, a minimum jerk trajectory is first plotted between the character locations. We then add two types of noise to the path. The first type adds noise to the starting and ending positions of the path. The $(x, y)$ coordinates for these positions are sampled from a 2D Gaussian distribution centred at the middle of the corresponding key on the keyboard. The second type adds noise to the path between the characters. To perform this, we first sample a coordinate pair $(x^\prime, y^\prime)$ from the uniform distribution of points bounded by the x and y coordinates of the starting and ending points of the path. Following this, we sample points on a path of minimum jerk passing through the 3 points uniformly over time \cite{wada2004via}. This process is repeated for every pair of adjacent characters in the word to obtain a sequence of coordinates that describes the trace. Figure ~\ref{fig:gesture_sample}(b) shows the trace created for the word \textit{budhi} using this method. Apart from the $(x, y)$ coordinate pair, we augment the input features for each sampled point with the $x$ and $y$ derivatives at the  point and a one-hot vector with value 1 at the index $i$  corresponding to the character on which the point lies on the keyboard. 


\section{Model}

\subsection{CTC Gesture Path Decoding}\label{sec:ctc_path_decoding}
\begin{figure}[!hbt]%
    \centering
    \subfloat{{\includegraphics[scale=0.14]{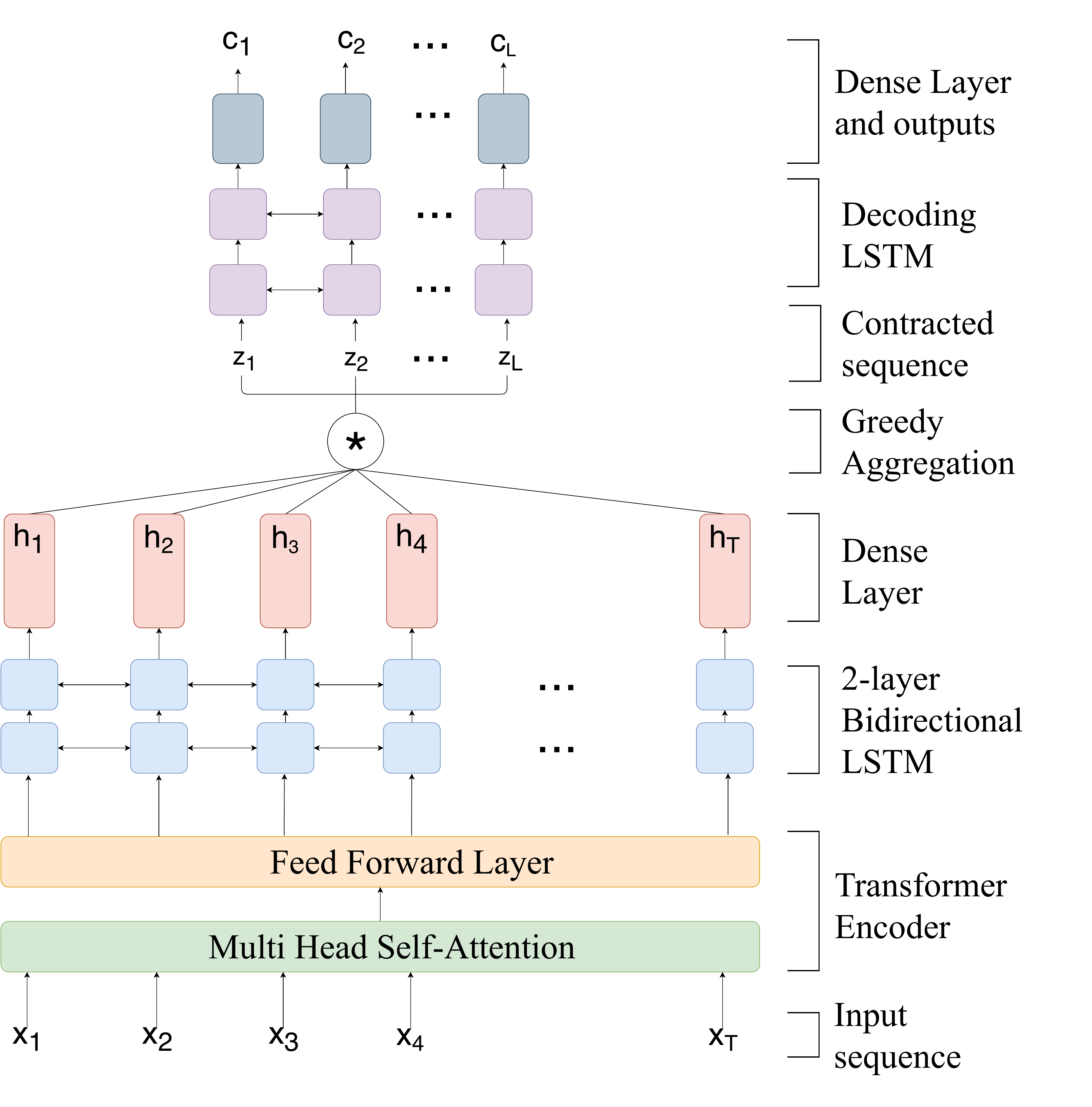} }}%
    \qquad
    \subfloat{{\includegraphics[scale=0.14]{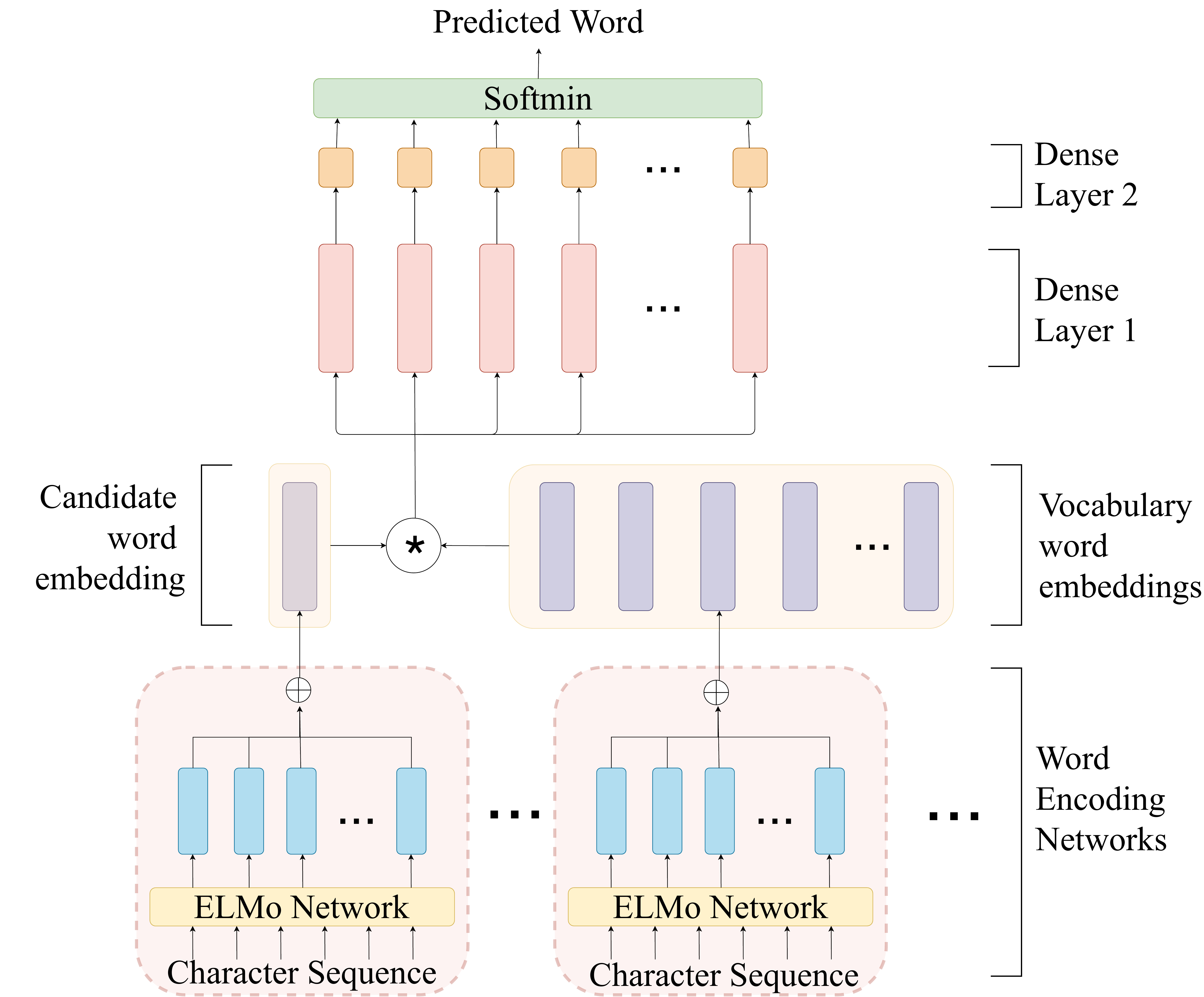} }}%
    \caption{(a) CTC Gesture Path Decoding Module Architecture consisting of the Transformer and LSTM encoder layers, greedy aggregation and character decoding LSTM layers; (b) Contrastive Transliteration Correction Architecture consisting of ELMO-based word encoding network, squared distance computation (shown by *) and fully connected layers.}%
    \label{fig:CTC_decoder_pd}%
\end{figure}

This module processes the input sequence of path coordinates to predict a sequence of English characters (which must be further transliterated into the Indic language) for the English-to-Indic decoding task and a sequence of Indic characters for the Indic-to-Indic decoding task. Consider an input sequence, $\{x_1, x_2, x_3, \ldots, x_T\}$ containing $(x, y)$ path coordinates along with augmented features described in Section \ref{sec: data_gen}. As seen in Figure ~\ref{fig:CTC_decoder_pd}(a), the sequence is passed into a Transformer encoder consisting of a multi-head self-attention sub-layer and a position-wise feed forward neural network, followed by a 2-layer Bidirectional LSTM network \cite{schuster1997bidirectional} to produce an encoded representation of the sequence. The encoded vector at each timestep is then passed through a fully connected layer with softmax activation to generate a sequence of vectors $\{h_1, h_2, h_3, \ldots, h_T\}$; where $h_i \in \mathbb{R}^{|C|+1} $, $h_{ij}$ is the probability of the $j^{th}$ character at timestep $i$ and $|C|+1$ is the number of characters including a blank character $<b>$. The model is trained using the CTC loss function which maximises the sum of probabilities of all frame-level alignments of the target sequence. Concretely, the CTC loss function maximises the probability:

\begin{equation}
    p(\textbf{y}|\textbf{x}) = \sum_{\hat{\textbf{y}}\epsilon \mathcal{A}_{ctc}(\textbf{y})}^{}p(\hat{\textbf{y}}|\textbf{x})
\end{equation}

\begin{equation}
    p(\hat{\textbf{y}}|\textbf{x}) = \prod_{t=1}^{T}p(\hat{y}_t|{\textbf{x}}) 
\end{equation}
where $\textbf{x}$ is the input sequence, $\textbf{y}$ is the target sequence of length $T$ and $\mathcal{A}_{ctc}(\textbf{y})$ is the set of all frame-level alignments of $\textbf{y}$. 
    
Unlike conventional CTC-based models, we do not use greedy or beam-search path decoding to infer the character sequence directly from $\{h_1, h_2, h_3, \ldots, h_T\}$. Instead, if all the vectors in a contiguous subsequence of length $k$ (say, $\{h_m, h_{m+1}, h_{m+2}, \ldots, h_{m+k-1}\}$) have the same most probable character (say, $c$), they are averaged to form a single vector $z_{m:m+k-1} \in \mathbb{R}^{|C|+1} $.

\begin{equation}
\begin{split}
    z_{m:m+k-1} = \frac{1}{k}\sum_{p=m}^{m+k-1}\{h_{p} | \arg \max_{j}h_{p,j} = c, c \neq \arg(<b>) \}
\end{split}
\end{equation} 

Following this, all the averaged vectors except for those corresponding to the blank character $<b>$ are concatenated into a new contracted sequence. We refer to this step as Greedy aggregation. Vectors corresponding to $<b>$ are used only to break the preceding contiguous subsequence and is particularly useful for modelling consecutive occurrences of the same character.  This sequence is passed into a 2-layer Bidirectional LSTM which models the co-character dependence within the word, followed by a fully connected layer with softmax activation. The output gives the final probability distribution over the characters and the most probable character is chosen at each timestep.

\subsection{Transliteration Generation}\label{sec:translit_gen}
\begin{figure}%
    \qquad
    \subfloat{{\includegraphics[scale=0.19]{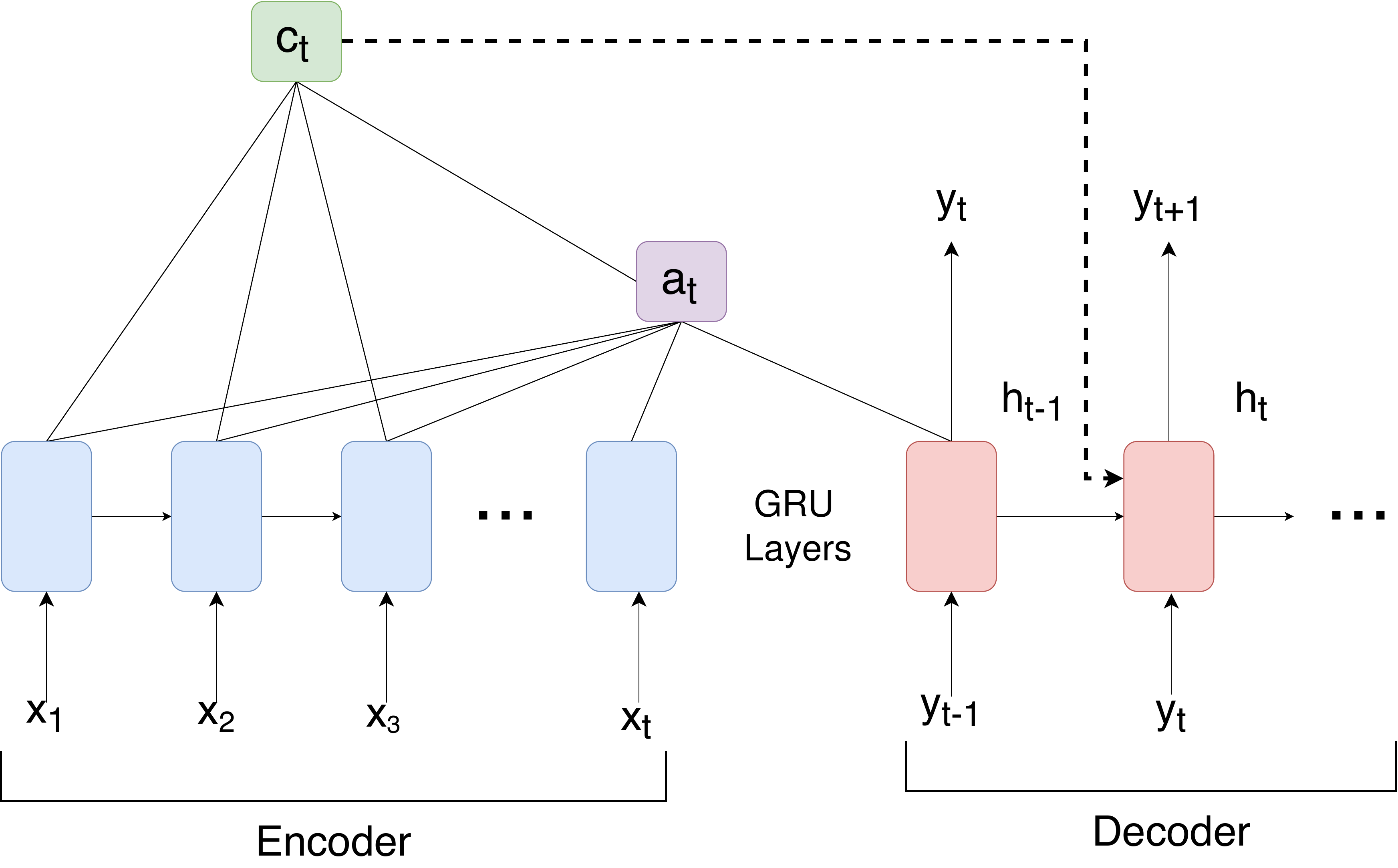} }}%
    \qquad
    \qquad
    \subfloat{{\includegraphics[scale=0.45]{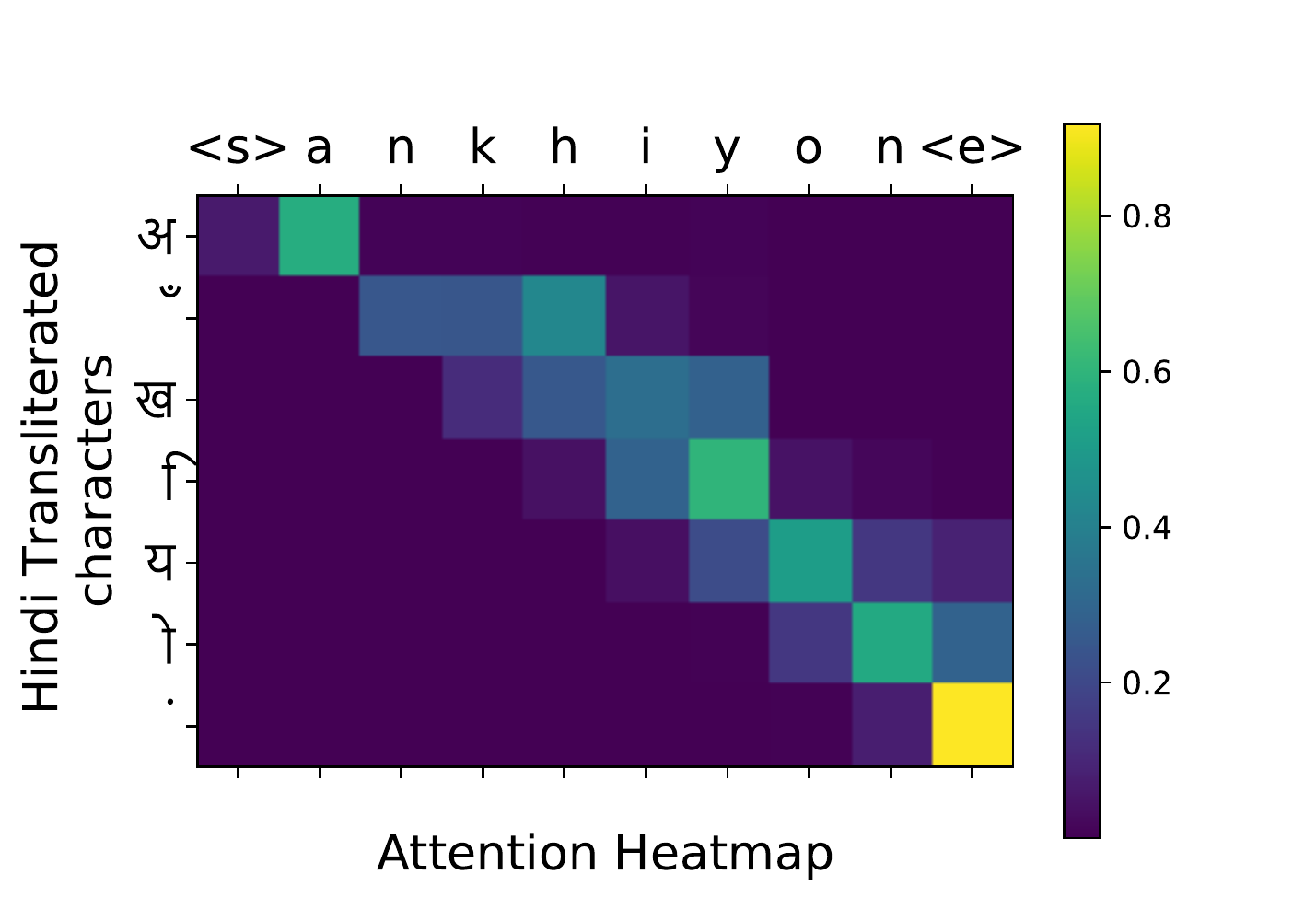} }}%
    \caption{(a) Transliteration Generation Architecture consisting of the GRU Encoder-Decoder framework with Bahdanau attention; (b) Heatmap of attention weights obtained during transliteration of the word \textit{`Ankhiyon'}  using the Transliteration Generation model.}%
    \label{fig:translit_attn}%
\end{figure}

This model is used only in the case of English-to-Indic decoding to transliterate the English character sequence generated by the CTC Gesture Path Decoder into the Indic language. The sequence is passed to a unidirectional GRU encoder to generate encoded representations corresponding to each character. A GRU decoder predicts the Indic characters corresponding to the transliterated word. At any time during decoding, the decoder uses its last hidden state and all the encoded vectors through Bahdanau attention to generate the next character as seen in Figure ~\ref{fig:translit_attn}(a). This module is trained on the character sequences generated by the CTC Gesture Path Decoder. We perform Beam Search Decoding on the GRU outputs to obtain the top 3 sequence predictions. A sample heatmap of attention weights generated by this module is shown in Figure ~\ref{fig:translit_attn}(b).

\subsection{Contrastive Transliteration correction}\label{sec:translit_corr}
This model is used to perform spelling correction on Indic words generated by the Transliteration Generation model in the English-to-Indic decoding task and by the CTC Gesture Path Decoder in the Indic-to-Indic decoding task. Word embeddings like GLoVe \cite{pennington2014glove} and Word2Vec \cite{mikolov2013distributed} generate vector representations of words that map semantic similarities to geometrical closeness. The ELMo \cite{peters-etal-2018-deep} network creates contextual word embeddings that depend on the entire sequence of words. It treats a sentence as a sequence of words and uses CNN and LSTM layers for feature extraction to obtain the word embeddings. 

In this work, we propose a method to create vector representations of words that are indicative of closeness in their character sequences by using the ELMo network for generating character embeddings by processing each word as a sequence of characters. The resulting embeddings would depend on other characters in the word and their relative positioning, thus making it suitable for spelling comparison. As seen in Figure ~\ref{fig:CTC_decoder_pd}(b), the generated character embeddings of a word are summed together and then min-max normalized  to obtain an encoding for the word. This process is followed to obtain an encoding for the misspelt word, $w$ and for each word $v$ in the vocabulary $\mathcal{V}$. Then, the model computes the squared difference between the encodings of the word $w$ and each of the words $v$ to obtain a distance vector, $d_{w, v_i} \in \mathbb{R}^{|E|}$, where $|E|$ is the ELMo embedding dimension. 
\begin{equation}
    d_{w, v} = (h_w - h_{v})^2; v \in |\mathcal{V}|
\end{equation}
where $h_w$ and $h_v$ represent the word encoding for the words $w$ and $v$ respectively. 

Distance vectors for each word pair are passed through two fully connected (dense) layers to obtain a 1-D distance metric, $e_{w, v}\in \mathcal{R}^1$. The word $v^*$ from the vocabulary for which distance metric, $e_{w, v^*}$ is the smallest is chosen as the corrected word. To accommodate words that are not part of the vocabulary, we place an upper threshold on $e_{w, v^*}$ and force the model to output the prediction from the Transliteration Generation module if it exceeds this threshold.
\begin{equation}
    v^* = \arg \min_{v \in |\mathcal{V}|} e_{w, v}
\end{equation}

\subsection{Implementation Details} \label{sec:eng_to_ind_sec}
\textbf{English-to-Indic decoding:} The transformer encoder in the CTC Gesture Path Decoder uses multi-headed attention with 5 heads, hidden layer size of 128, dropout rate of 0.05 and ReLU activation in the Attention and Feed-Forward layers. The hidden layer sizes of all Bidirectional LSTM layers in this module are fixed to 256. This module is trained using the CTC loss function with a learning rate of 0.01 over 20 epochs. The Transliteration Generation module uses a single Unidirectional GRU layer with a hidden size of 512. This module is trained over 10 epochs using the Categorical Cross-Entropy loss function and Beam Search decoding with a beam size of 3 is performed on the GRU outputs. The 3 predictions are passed as independent samples to the Contrastive Transliteration Correction module for generating 3 suggestions for the final word, which may not all be unique. The Contrastive Transliteration Correction module has hidden layers of size 64 and 1. It is trained using the Sparse Categorical Cross-Entropy loss function. The Adam optimizer is used for training all the three modules.  

\textbf{Indic-to-Indic decoding: }In this case, we make a few architectural changes to the model. Since the CTC Gesture Path Decoder directly predicts an Indic character sequence, we remove the Transliteration Generation module from the pipeline. Besides, we reduce the inference time and model complexity by removing the two Bidirectional LSTM layers of the CTC Gesture Path Decoder. We cannot afford to do this in English-to-Indic decoding as we require more accurate predictions to prevent compounding of errors as it passes through the Transliteration Generation module. The model parameters for the rest of the architecture remain the same as described in section \ref{sec:eng_to_ind_sec}. For both the tasks, a train-validation-test split of 70-20-10 is used for all languages.
  
\section{Results and Discussion}
    
In this section, we analyse our model's performance on both tasks across 7 different Indic languages, namely Hindi, Bengali, Gujarati, Tamil, Telugu, Kannada and Malayalam. These languages together account for the native languages of 75\% of the Indian population. The first three languages are Indo-Aryan languages and the last four are Dravidian languages. Therefore, the languages have distinct lexical, syntactic and grammatical properties, which makes the task of creating a generalizable model more challenging and worthwhile. In Table~\ref{tab:eng2ind}, we present the results of our model on the English-to-Indic Decoding task. The final accuracy of the model is defined as the percentage of input traces that are correctly processed by the model to generate the Indic word. When the beam size $k$ is greater than 1, the model is said to make a correct prediction if at least one of the $k$ predictions is accurate. We also present the intermediate accuracies observed after each module in our architectural pipeline and the accuracies obtained with various beam sizes.

\begin{table*}[!hbt]
  \centering
  \renewcommand{\arraystretch}{0.9}
  \begin{tabular}{|p{2.2cm}|c|c|c|c|c|c|c|c|}
    \hline
    \thead{Indic\\Language} & \thead{Dataset\\Size} & \thead{CTC Path\\Decoding (\%)} & \multicolumn{3}{c|}{\thead{Transliteration\\Generation Acc. (\%)}} & \multicolumn{3}{c|}{\thead{Transliteration\\Correction Acc. (\%)}} \\\cline{4-9}
    & & & \textbf{k = 1} & \textbf{k = 2} & \textbf{k = 3} & \textbf{k = 1} & \textbf{k = 2} & \textbf{k = 3 (Final)}\\
    \hline
    Hindi & 29074 & 98.45 & 59.14 &	73.12 & 77.56 & 71.53 & 83.86 & \textbf{89.12}  \\ 
    Tamil & 21523 & 98.02  & 53.31 & 60.98 & 65.47 &   71.10 & 80.47 & \textbf{86.96}  \\ 
    Bengali & 17332 & 98.60 & 23.07 & 36.01 & 43.40 &   63.02 & 73.94 & \textbf{81.13}  \\ 
    Telugu & 12733 & 96.64 & 21.28 & 34.81 & 38.19 &   57.39 & 67.68 & \textbf{78.21}  \\ 
    Kannada & 4877 & 96.95 & 22.13&	28.83  & 37.60 &   47.75 & 64.98 & \textbf{72.00}  \\ 
    Gujrati & 8776 & 98.78 & 17.99 &	28.57  & 38.12 &   53.12 & 60.05 & \textbf{72.65}  \\ 
    Malayalam & 10097 & 98.21 & 18.83 & 27.57 & 31.61 &  48.32 & 60.41 & \textbf{70.00}  \\ \hline
  \end{tabular}
  \caption{Results on English-to-Indic Decoding across 7 Indic languages}\label{tab:eng2ind}
\end{table*}

In Table~\ref{tab:indtoind}, we present the final accuracy of our model on the Indic-to-Indic Decoding task, along with the intermediate accuracy after the CTC Gesture Path Decoder module. 

\begin{table*}[!thb]
\centering
    \begin{tabular}{|p{2.6cm}|c|c|c|}
    \hline
      \thead{Indic Language} & \thead{Dataset Size} & \thead{CTC Accuracy (\%)} & \thead{Final Accuracy (\%)} \\ \hline
     Hindi &  {32415}  & 68.93 & \textbf{86.75}  \\ 
     Tamil &  {24094}  & 68.16 & \textbf{95.73}  \\ 
     Bengali & {15320}  & 75.06 & \textbf{88.74}  \\ 
     Telugu &  {28996}  & 74.94 & \textbf{94.86}  \\ 
     Kannada &  {26551}  & 66.78 & \textbf{89.06}  \\ 
     Gujarati &  {30024}  & 61.21 & \textbf{82.13}  \\ 
     Malayalam &  {36258}  & 56.98 & \textbf{95.53}  \\ \hline
    \end{tabular}
    \caption{Results on Indic-to-Indic Decoding across 7 Indic languages}
    \label{tab:indtoind}
    \end{table*}
    
We observe in Table \ref{tab:eng2ind} that while the accuracy of the CTC Gesture Path Decoder is not affected by the decrease in dataset size, the transliteration accuracy shows a decreasing trend. This is mainly because the transliteration generation module requires a large dataset to learn phonological relationships. The accuracies in Table \ref{tab:indtoind} are not highly correlated with the dataset size due to the absence of this module in the corresponding architecture. Another factor that influences accuracy is the word length as longer words are less likely to be correctly decoded by the Transliteration Generation module. On the contrary, we observed that the performance of the Transliteration Correction module is not much affected by word lengths as the 1024-dimensional ELMo character embedding allows each character to have a distinct influence on each dimension in the summed encoding. This allows even longer character sequences to be compared by this module. The combined effect of word lengths on the two modules explains a greater percentage improvement in accuracy after correction for certain languages which have greater average word lengths in our dataset like Malayalam and Telugu (word lengths of 12.39 and 9.02 respectively) as opposed to others like Hindi and Bengali (6.55 and 6.95 respectively).

Besides,we observe that the CTC Gesture Path Decoder accuracies are higher for  English-to-Indic Decoding than Indic-to-Indic Decoding. This is because (i) the module uses two additional Bidirectional LSTM layers that process the outputs of the Transformer encoder, resulting in greater representative power and (ii) Indic scripts contain a larger character set than English resulting in a denser keyboard with greater chances of error. Nevertheless, we find that the final accuracies for the Indic-to-Indic decoding task after spelling correction using the Transliteration Correction module are high which highlights the importance of this module. It also means that the character sequences predicted by the CTC Gesture Path Decoder closely match the expected sequence due to which the nearest word from the vocabulary match the expected word. 

In the following sections, we analyse the performance of certain architectural variants and also perform a detailed error analysis of each module in our architecture. Throughout the following study, we present the results on the English-to-Indic Decoding task, with Hindi as the Indic language unless otherwise specified.

\subsection{Performance Comparison} 
In this section, we compare the performance of our model with alternative architectures and existing work. In Table ~\ref{tab:spell_check_variants}, we first observe that the CTC Gesture Path Decoder of our model achieves a higher accuracy than that of the \textbf{CTC-LSTM} model \cite{alsharif2015long} which does not use a transformer encoder or $x$ and $y$ derivative values in the feature vectors. Next, we compare the accuracy of the transliteration component of our architecture comprising the Transliteration Generation and Correction modules against the results obtained by performing transliteration using (i) a \textbf{Transformer}-based Encoder-Decoder model; (ii) \textbf{DeepTranslit}\footnote{\url{https://github.com/notAI-tech/DeepTranslit}}; (iii) an LSTM-based \textbf{Seq2Seq model}; (iv) an LSTM-based Encoder-Decoder framework with \textbf{Bahdanau Attention}; (v) a \textbf{rule-based model} \cite{lavanya2005simple} and (vi) a \textbf{BiLSTM-CTC model} \cite{rosca2016sequence}. Beam Search Decoding with $k$ = 3 is performed on the outputs of models (i)-(iv) before computing the accuracy.

Transliteration is highly reliant on the sequential nature of the input and in most cases, requires modelling of short-range and past-input dependencies. This explains why GRU and LSTM based models outperform the transformer-based model which attempts to also learn long-range dependencies on past and future inputs. Besides, the better accuracy of our Transliteration Generation module which uses GRUs over similar LSTM-based models can be attributed to LSTMs modelling longer-term dependencies and GRUs usually performing better on less training data. While the CTC loss function has shown good results in gesture path decoding, the same does not hold for the transliteration task as shown by the BiLSTM-CTC model which only gives an accuracy of 47.96\%. 

\begin{table}[!bht]
   \subfloat{\begin{tabular}{| c | c |}
      \hline
       \thead{Gesture Path Decoding Variant} & \thead{Acc. (\%)} \\ \hline 
     CTC-LSTM \cite{alsharif2015long} & {95.23} \\
      \textbf{Our Model: CTC Gesture Path Decoder} & \textbf{98.45}\\ \hline 
      
      \thead{Transliteration Variant \\ (Generation)} & \thead{Acc. (\%)} \\ \hline
      Transformer &  {37.51}  \\ 
      DeepTranslit  & {60.90}  \\ 
      Seq2Seq (LSTM) & {75.71} \\  
      Bahdanau Attention (LSTM) & {75.87}   \\  
      Rule-based \cite{lavanya2005simple} & {20.85} \\  
      BiLSTM-CTC \cite{rosca2016sequence} & 47.96 \\ \hline
       \end{tabular}}%
      \qquad
   \subfloat{\begin{tabular}{ | c | c | }
      \hline
      \thead{Transliteration Variant\\(Correction)} & \thead{Acc. (\%)}\\ \hline
      Average ELMo  &  {81.11}  \\ 
      Weighted Average ELMo  &  {81.05}  \\ 
      Unnormalized ELMo  & {76.65}   \\
      Word2Vec Embeddings & {68.99} \\
      NPLM \cite{bengio2003neural} &  {82.25}  \\
      Cosine Similarity & {78.63}  \\ 
      Reordered Model & {77.23}\\ \hline
       \thead{Our Model: Transliteration \\Generation + Correction} &  {\textbf{89.12}}  \\ \hline
    \end{tabular}}%
    \caption{Comparison of model performance with various architectural variants}\label{tab:spell_check_variants}
\end{table}

We also present the results obtained on variants of our Contrastive Transliteration Correction module which uses min-max normalized ELMo embeddings. For this study, we pass the outputs generated by our proposed Transliteration Generation module after Beam Search Decoding ($k$ = 3) to the following variants of our Correction module. \textbf{Average ELMo} takes the average of the ELMo vectors corresponding to the 3 beam search outputs and passes it as a single input. This reduces computational requirements but brings the accuracy down to 81.11\% due to mixing of information from the 3 decoded outputs. \textbf{Weighted Average ELMo} performs weighted averaging of the ELMo vectors (with weights 0.5, 0.3, 0.2 for the 3 beam search outputs) but fails to show improvement in accuracy. \textbf{Unnormalized ELMo} does not perform min-max normalization of the word encoding which is important to remove bias due to variation in the word lengths. \textbf{Word2Vec Embeddings} uses Word2Vec word embeddings instead of the summed ELMo character embeddings which results in the accuracy dropping to 68.99\%. Word embeddings do not encode the relative positioning of characters and are therefore, less representative of the character sequence. \textbf{NPLM} uses the Neural Probabilistic Language Model \cite{bengio2003neural} instead of ELMo to generate the character embeddings. The observed fall in accuracy to 82.25\% is because ELMo character embeddings are able to better model differences in character sequences. 

\textbf{Cosine Similarity ELMo} is an architectural variant which chooses the word whose encoding has the highest cosine similarity with the encoding of the candidate character sequence. Since, the cosine similarity score is a scalar value, the dense layers in the Correction module architecture are removed in this case. The fall in accuracy from 89.12\% (for our model) to 78.63\% shows that augmenting the euclidean distance vectors using dense layers gives a better measure of the closeness between the encodings. The same trend was observed in other languages as well. For example, this variant reduced the accuracy from 86.96\% to 80.13\% for Tamil and from 70.00\% to 64.23\% for Malayalam. The \textbf{Reordered Model} considers a reordering of our architecture that places the Contrastive Transliteration Correction module immediately after CTC Gesture Path decoder and followed by the Transliteration Generation module. The Correction module in this case corrects errors in the CTC output and uses a vocabulary of correct English transliterations of Indic words. As is evident from Table \ref{tab:eng2ind}, much of the prediction error in our pipeline is induced by the Transliteration Generation module, due to which the Transliteration Correction module in this altered position is much less rewarding. 

Finally, we modify the Indic-to-Indic decoding architecture to study the effect of including the two bidirectional LSTM layers (as in English-to-Indic decoding). We observe that while the average CTC Gesture Path Decoder accuracy across the 7 languages rises from 67.43\% (for our model) to 89.5\%, the average final accuracy only rises from 90.4\% to 94.5\%. Besides, we observe a 64\% increase in  training time and 75\% increase in inference time for the modified architecture. Thus, the gains from the additional layers are small.

\subsection{Error Analysis}
\begin{figure}[!thb] 
\centering
    \subfloat{{\includegraphics[scale=0.32]{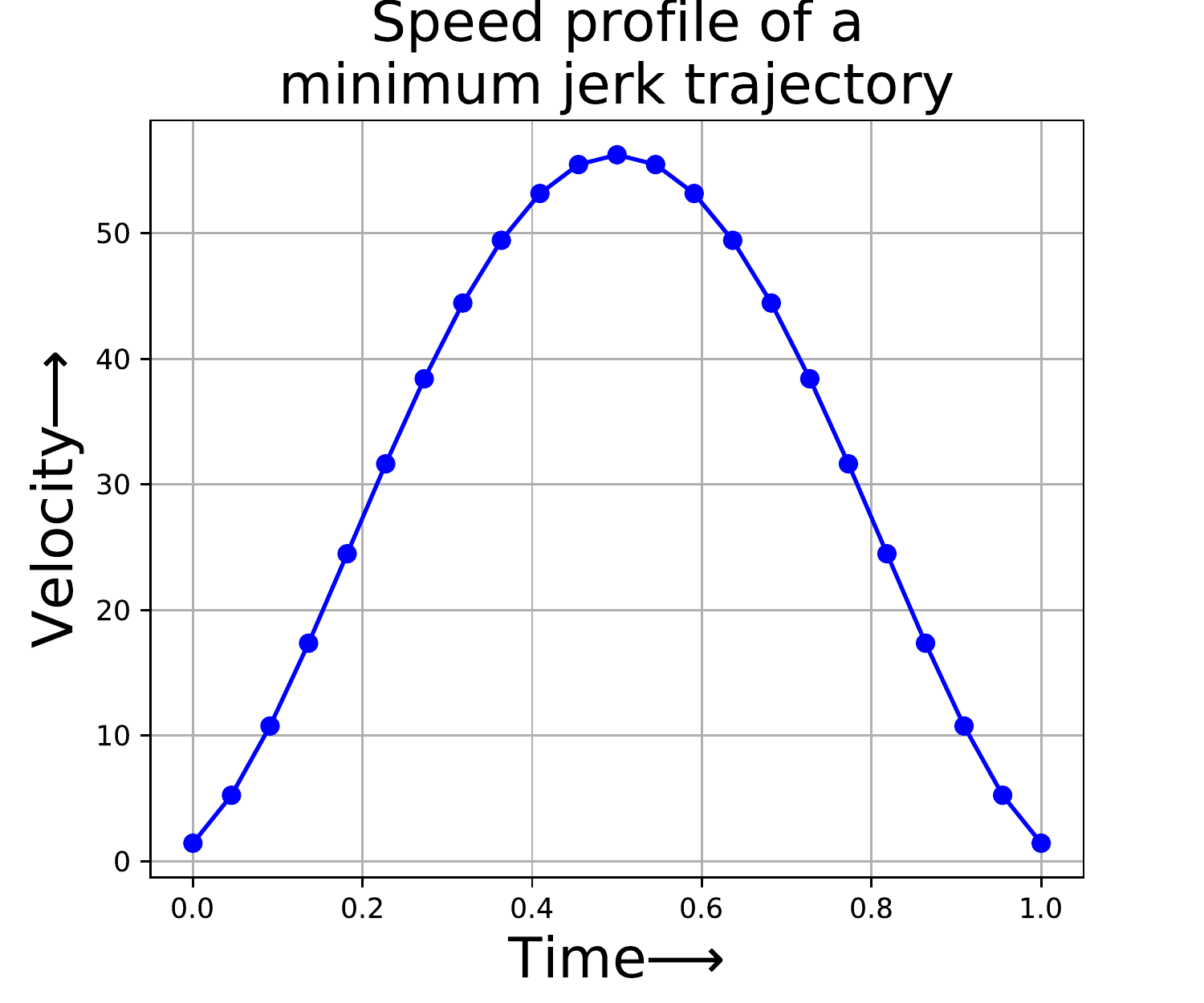}}}%
    \qquad
    \subfloat{{\includegraphics[scale=0.32]{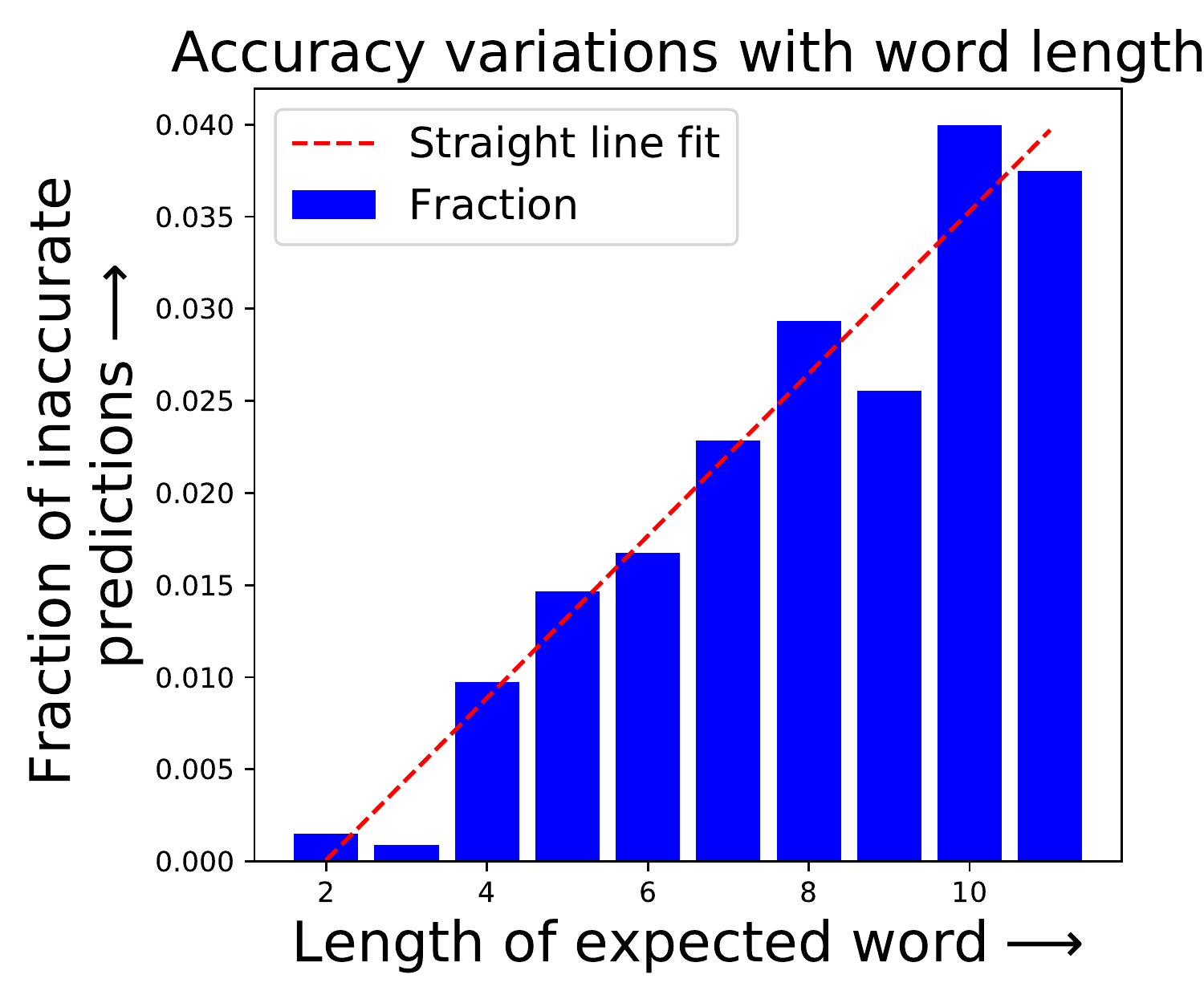}}}%
    \caption{(a) Speed profile of a minimum jerk trajectory; (b) Accuracy variation with length of word} 
    \label{fig:analysis_1}%
\end{figure}

The speed profile of a minimum jerk trajectory has a minimum at the endpoints and reaches the maximum value midway between the two points as seen in Figure \ref{fig:analysis_1}(a). Therefore, when points are sampled on the path at equal time intervals, there is less relative distance between points near the locations of the intended characters. The CTC Gesture Path Decoder is found to make more confident predictions in cases where the corresponding keys are farther apart on the keyboard as the variations in relative distances between points are better emphasised. For example, the model wrongly predicts \textit{mashhuur} and \textit{kaare} as \textit{mashuur} and \textit{kare} where consecutive characters at the same position are missed. Besides, the fraction of words that are wrongly predicted by the model increases with the word length as seen in Figure ~\ref{fig:analysis_1}(b).

\begin{figure}[!hbt]%
    \subfloat{{\includegraphics[scale=0.34]{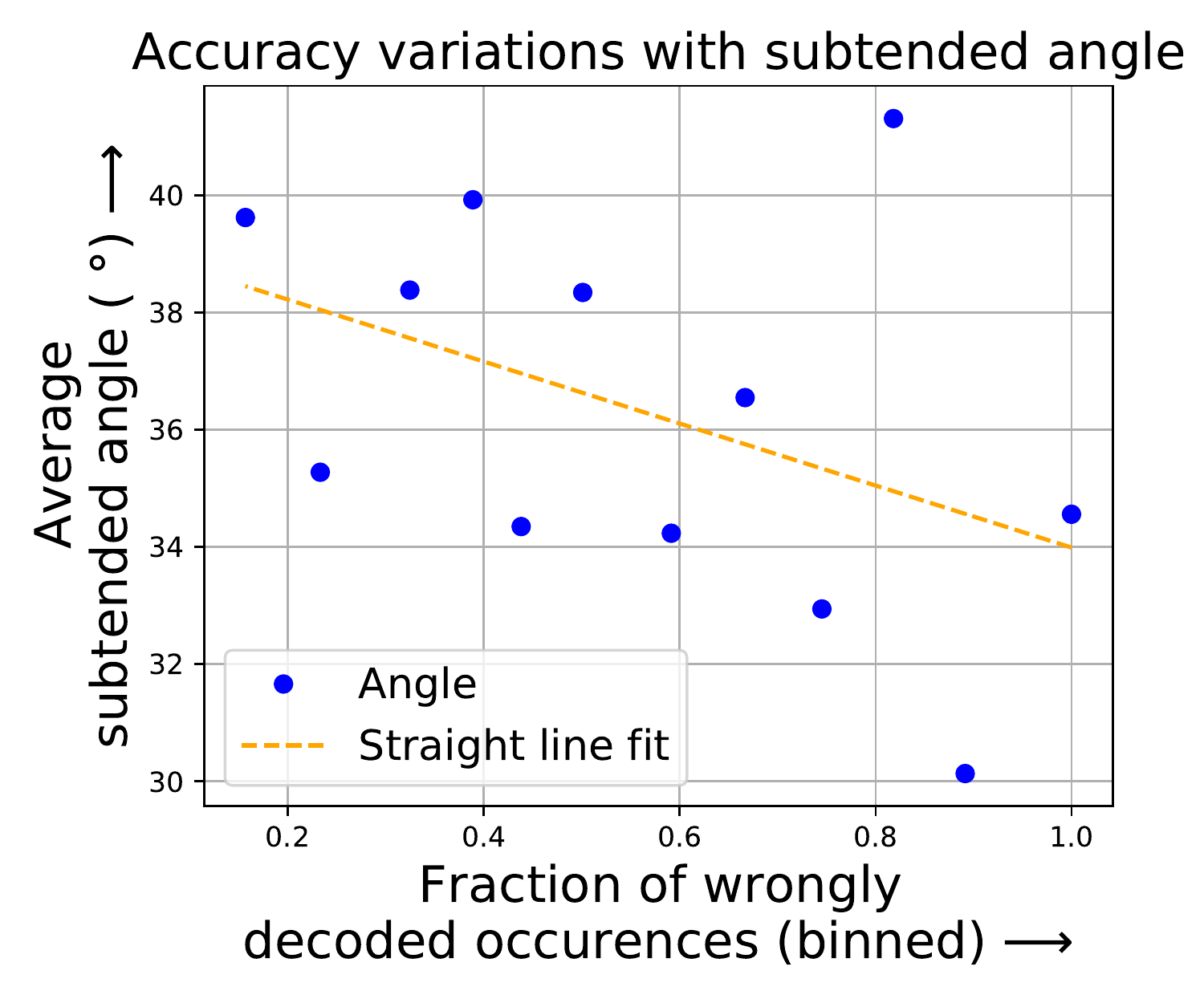} }}%
    \subfloat{{\includegraphics[scale=0.34]{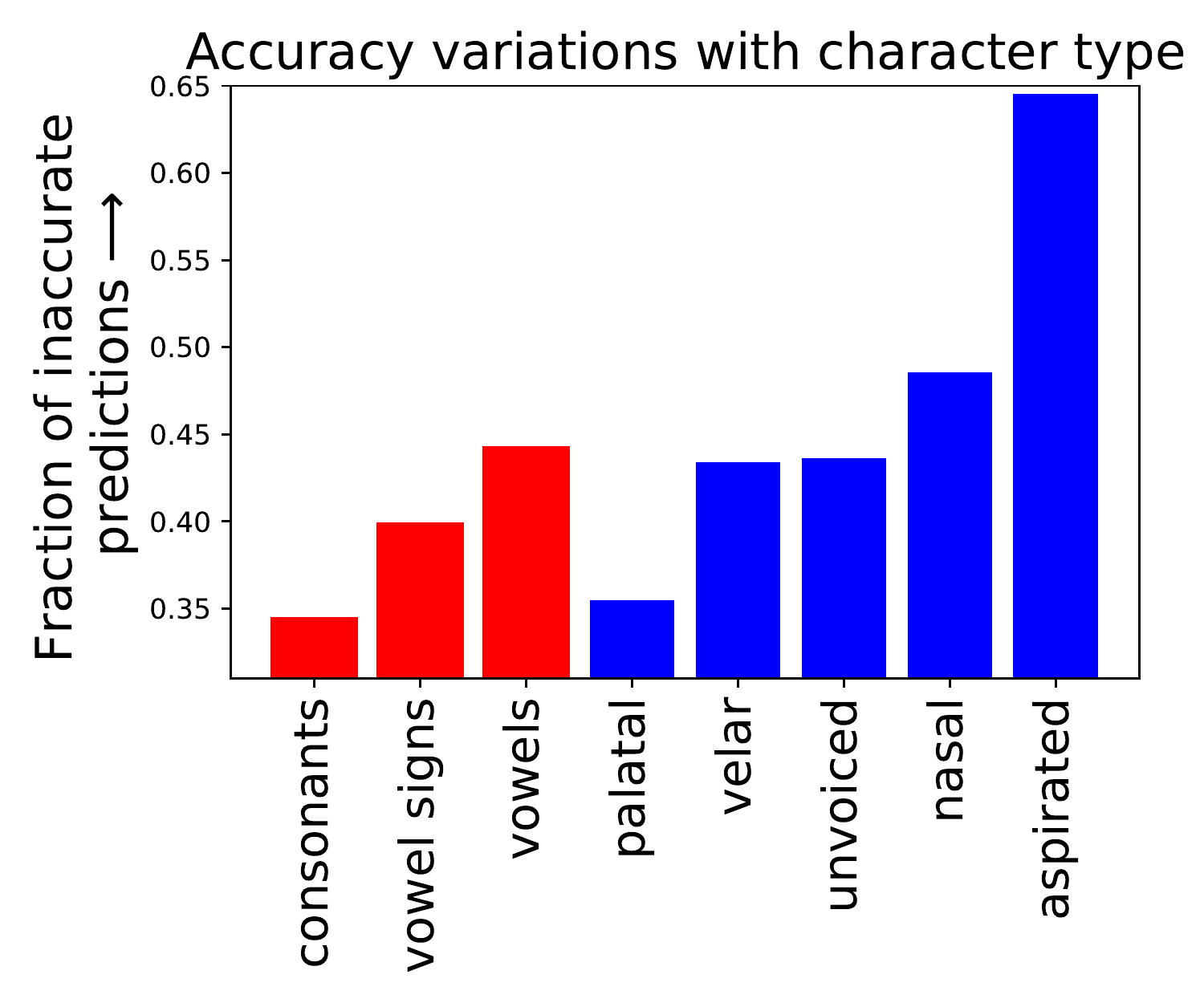}}}%
    \subfloat{{\includegraphics[scale=0.34]{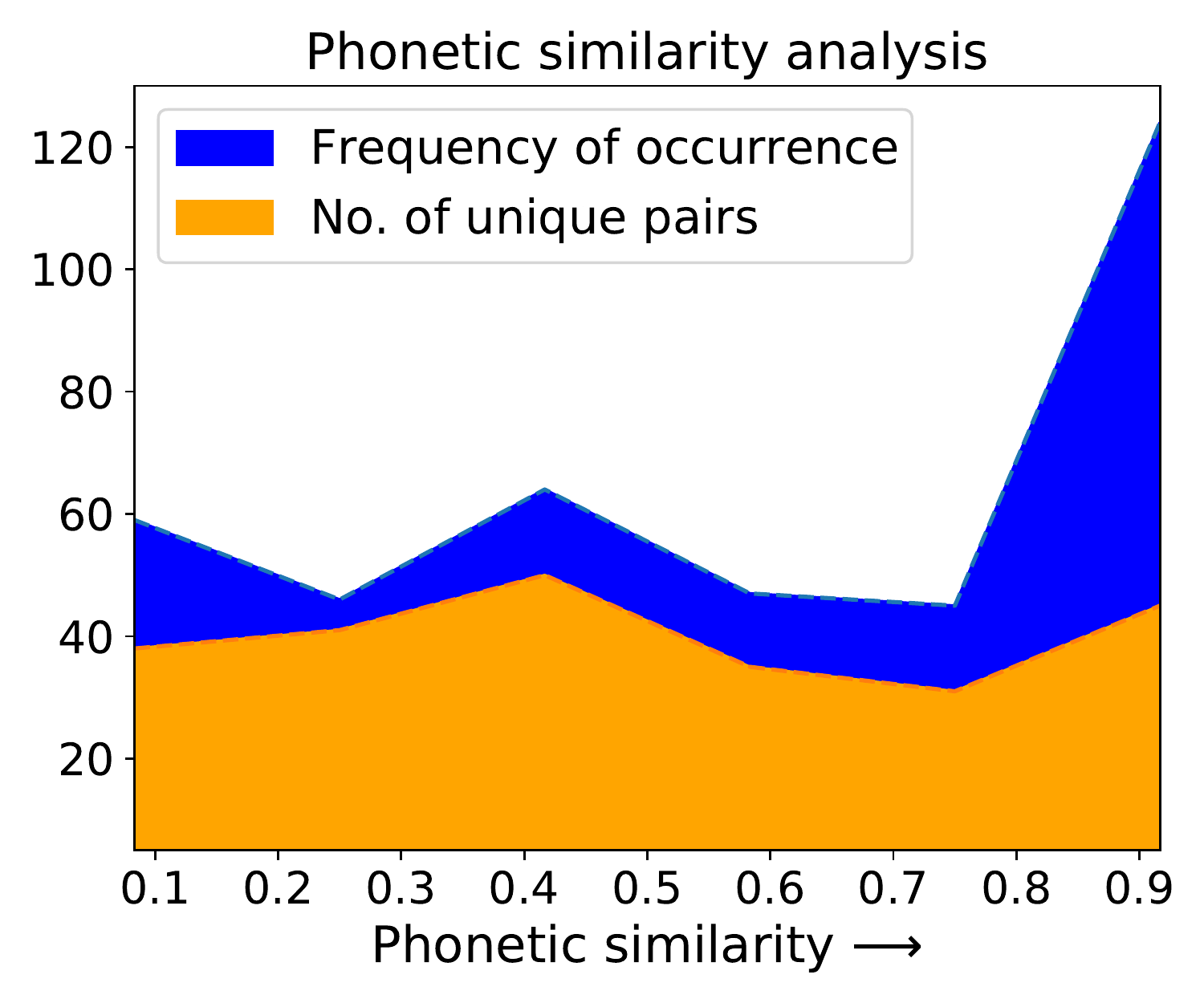} }}%
    \caption{(a) Accuracy Variation with Subtended Angle; (b) Accuracy variations with character type; (c) Phonetic Similarity analysis}%
    \label{fig:analysis_plots}%
\end{figure}

The CTC Gesture Path Decoder also depends on the directional information from the $x$ and $y$ derivative values at each sampled point on the path. Therefore, a few of the words with more than two consecutive characters occurring on the same horizontal line on the keyboard were wrongly decoded. For example, the word \textit{badhoge} has the characters \textit{a}, \textit{h} and \textit{g} on the same keyboard row and is wrongly predicted as \textit{bahoge}. We extracted all 3-grams of characters from words used in our test set and grouped them into bins based on their empirical probability of being wrongly decoded. We then computed the acute angle subtended by the path connecting the characters of each 3-gram and averaged the angles within each bin. From Figure ~\ref{fig:analysis_plots}(a), we see that the accuracy reduces as the path subtends a smaller angle.

To analyse errors in transliteration, we classify the Hindi character set based on (i) the character type into vowels, vowel signs and consonants, and (ii) the phonetic sound into palatal, velar, unvoiced, nasal and aspirated (as seen in Table ~\ref{tab:Grouping of characters in Hindi}(a)). Figure ~\ref{fig:analysis_plots}(b) shows the fraction of characters within each character type that are wrongly predicted. We observe that aspirated characters are more error-prone which is primarily because the English source scripts do not clearly distinguish them from their unaspirated counterparts. For example, the model wrongly predicts the aspirated character {\dn T} in {\dn TAro} as {\dn t} resulting in {\dn tAro}. Inaccurate predictions in nasal characters are also due to ambiguities in the English source script. For example, \textit{`an'} is used to refer to both the characters {\dn n} and {\dn Z} as seen in {\dn bQcn} (\textit{bachchan}) and {\dn vAtAvrZ}  (\textit{vaatavaran}). Errors in predicting vowels are caused due to the model sometimes predicting their respective vowel signs when they occur in between two characters in a word. 

In Figure ~\ref{fig:analysis_plots}(c), we consider all words that differ from the predictions corresponding to them by a single character and compute the phonetic similarity between the expected and wrongly predicted characters. A large number of character pairs having a high phonetic similarity shows that the wrong predictions are close to the expected words. As the Contrastive Transliteration Correction module relies on proximities in word spellings, most of the inaccuracies are observed in cases where several words in the vocabulary have spellings similar to the character sequence they are compared with. 

\begin{table}[!thb]%
\subfloat{\begin{tabular}{ | c | c |}
      \hline
    \thead{Phonetic Type} & \thead{Character list} \\ \hline
      Unvoiced &  {\dn{k}$,$ \dn{t}$,$ \dn{c}$,$ \dn{p}$,$ \dn{T}$,$ \dn{V}$,$ \dn{K}$,$ \dn{C}$,$ \dn{W}$,$ \dn{P}} \\ 
      Aspirated & {\dn{T}$,$ \dn{J}$,$ \dn{K}$,$ \dn{C}$,$ \dn{B}$,$ \dn{G}$,$ \dn{W}$,$ \dn{P}$,$ \dn{D}$,$ \dn{X}}\\  
      Nasal  & {\dn{n}$,$ \dn{Z}$,$ \dn{B}} \\  
      Palatal &{\dn{c}$,$ \dn{C}$,$ \dn{j}$,$ \dn{J}}\\ 
      Velar & {\dn{k}$,$ \dn{K}$,$ \dn{g}$,$ \dn{G}}   \\  \hline
       \end{tabular}}%
       \qquad
\subfloat{\begin{tabular}{|>{\centering\arraybackslash}p{4.2cm}|>{\centering\arraybackslash}p{1.5cm}|}
  \hline
  \centering
      \thead{Ablated Component} & \thead{Acc. (\%)} \\ \hline
      Derivative features (CTC) &  {90.57}  \\ 
      Attention (Transliteration) & 75.71  \\  
      Transliteration Correction & 77.56 \\
      Dense (Transliteration \newline Correction) & 80.51 \\ \hline
    \end{tabular}}%
   \caption{(a) Phonetic grouping of Hindi characters;  (b) Results of the Ablation Study}
    \label{tab:Grouping of characters in Hindi}
\end{table}

\subsection{Ablation Study}
In this section, we analyse our architecture by performing an ablation study as reported in Table \ref{tab:Grouping of characters in Hindi}(b) on the following components to understand their relative importance: (i) \textbf{Derivative features (CTC)}: We remove the x and y derivative values for each sampled point from the input features to the CTC Gesture Path Decoder which provide information about the direction and speed of motion. This results in the path decoding accuracy falling from 98.45\% to 90.57\%. (ii) \textbf{Attention (Transliteration)}: We remove the Bahdanau attention mechanism from the Transliteration Generation module and this results in a fall in accuracy from 77.56\% to 75.71\%, mainly due to long words being wrongly decoded. (iii) \textbf{Transliteration Correction Module}: We remove the Contrastive Transliteration correction module and use the outputs of the Transliteration Generation module as the final prediction which gives a reduced accuracy of 77.56\%. (iv) \textbf{Dense (Transliteration Correction)}: We remove the dense layers in the Contrastive Transliteration correction module and use the Euclidean distance between the word encodings as a direct measure of their closeness. This reduces the accuracy from 89.12\% to 80.51\%. An analysis on other languages revealed similar trends with accuracy falling from 86.96\% to 80.51\% for Tamil and from 70.00\% to 63.50\% for Malayalam.

\section{Conclusion and Future Work}
In this work, we have developed an architecture for gesture input decoding on keyboards supporting Indic languages and validated our models across 7 languages. We have shown that a joint Transformer/RNN architecture that uses the CTC loss function can decode a gesture input with high accuracy. We have also shown that the character-level ELMo network can be used to generate spelling-aware word encodings and perform spelling correction efficiently. Going further, we wish to extend our work to support visually impaired users and diversify our dataset to a larger set of languages.

\bibliographystyle{coling}
\bibliography{coling2020}

\end{document}